%% file: main.tex
\documentclass[letterpaper, 10 pt, conference]{ieeeconf}  

\IEEEoverridecommandlockouts                              

\title{\LARGE \bf
Deep Imitation Learning for Humanoid Loco-manipulation\\through Human Teleoperation}

\author{Mingyo Seo, Steve Han, Kyutae Sim, Seung Hyeon Bang, Carlos Gonzalez, Luis Sentis, and Yuke Zhu
\thanks{The University of Texas at Austin.}
\thanks{Correspondence: {\tt\small mingyo@utexas.edu}}%
}

\let\labelindent\relax
\DeclareUnicodeCharacter{0301}{\'{e}}
\usepackage{times}
\usepackage{xcolor}

\usepackage{amsfonts}
\usepackage{amsmath}
\usepackage{amssymb}

\usepackage{graphicx}
\usepackage{svg}

\usepackage{booktabs}
\usepackage{hyperref}
\usepackage{dsfont}

\usepackage[font=small,labelfont=bf]{caption}
\usepackage{enumitem}
\usepackage[backend=biber,
            hyperref=true,
            url=false,
            isbn=false,
            doi=false,
            backref=false,
            style=ieee,
            citestyle=numeric-comp,
            sorting=nyt,
            block=none]{biblatex}
\usepackage[font=footnotesize]{caption}

\usepackage{algpseudocode}
\usepackage{algorithm}

\setlist[itemize]{leftmargin=*}

\renewcommand{\bibfont}{\small}
\newcommand{\ourmethod}{{\textsc{TRILL}}}

\newcommand{\better}[1]{{\tiny\textcolor{blue}{$\blacktriangle$#1}}}
\newcommand{\worse}[1]{{\tiny\textcolor{red}{$\blacktriangledown$#1}}}
\newcommand{\same}[1]{{\tiny{\quad\,\;\,\;}}}
\newcommand{\task}[1]{\textbf{\textit{#1}}}

\begin{document}

\maketitle
\thispagestyle{empty}
\pagestyle{empty}

%


\maketitle


\begin{abstract}
We tackle the problem of developing humanoid loco-manipulation skills with deep imitation learning. The difficulty of collecting task demonstrations and training policies for humanoids with a high degree of freedom presents substantial challenges.
We introduce \ourmethod{}, a data-efficient framework for training humanoid loco-manipulation policies from human demonstrations. 
In this framework, we collect human demonstration data through an intuitive Virtual Reality (VR) interface.
We employ the whole-body control formulation to transform task-space commands by human operators into the robot's joint-torque actuation while stabilizing its dynamics.
By employing high-level action abstractions tailored for humanoid loco-manipulation, our method can efficiently learn complex sensorimotor skills.
We demonstrate the effectiveness of \ourmethod{} in simulation and on a real-world robot for performing various loco-manipulation tasks. 
Videos and additional materials can be found on the project page: \url{https://ut-austin-rpl.github.io/TRILL}.

\end{abstract}

\input{introduction}

\input{related_work}
\input{method}
\input{experiments}
\input{conclusion}




{\small
\noindent
{\bf Acknowledgements}
We would like to thank Huihan Liu, Zhenyu Jiang, and Yifeng Zhu for providing feedback on this manuscript.
We acknowledge the support of the Office of Naval Research (N00014-22-1-2204).
}
\renewcommand*{\bibfont}{\footnotesize}
\printbibliography

\input{appendix}

\end{document}

%% file: introduction.tex
\section{Introduction}


Recent years have witnessed significant progress in hardware designs and control algorithms tailored for humanoid robots~\cite{saeedvand2019comprehensive, sugihara2020survey}.
Owing to their human-like embodiment, these robots possess enormous versatility for performing various everyday tasks in human-centric environments~\cite{darvish2023teleoperation}. 
However, the lack of autonomy presents a major obstacle to the widespread deployment of humanoids in the real world.
To date, most operational approaches for these robots heavily depend on task-specific manual programming~\cite{jorgensen2020finding, murooka2021humanoid, settimi2016motion} or human teleoperation~\cite{jorgensen2019deploying, ishiguro2020bilateral, penco2019multimode}.

Imitation learning has recently emerged as a flexible, data-driven approach for building robot controllers from human demonstrations~\cite{xie2020deep, zhang2018deep}. 
Particularly, deep imitation learning algorithms, implemented with large neural networks, have been successfully applied to simpler robot morphologies, including tabletop arms and wheeled platforms~\cite{mandlekar2021matters, brohan2022rt}.
However, applying these algorithms to humanoid robots presents two additional challenges.
The first challenge stems from the fact that humanoid robots are floating-base systems that need to maintain balance while physically interacting with the environment. 
In contact-rich tasks, the robot's physical interactions affect its dynamics and increase the uncertainty and complexity of controlling the robot.
This problem is further exacerbated by the absence of tactile and proprioceptive sensing modalities in standard teleoperation interfaces.
The second challenge lies in the humanoid robots' high degree of freedom, leading to a large action space that raises the data requirements and computational demands for policy learning.

Our key idea to overcome these challenges is incorporating the whole-body control formulation~\cite{sentis2006whole} into our data collection system and policy learning method. Whole-body control is a comprehensive control framework that employs a minimal set of simple, low-dimensional rules to fully leverage the capabilities of floating-based robots for compliant multi-contact interaction with their environment. Utilizing this controller simplifies the process for human operators to provide task demonstrations through an intuitive Virtual Reality (VR) interface. Moreover, it enables our policy to predict high-level actions in the task space. These actions can subsequently be transformed into joint-level torque commands for actuation.



\begin{figure}
	\centering
	\includegraphics[width=0.9\linewidth]{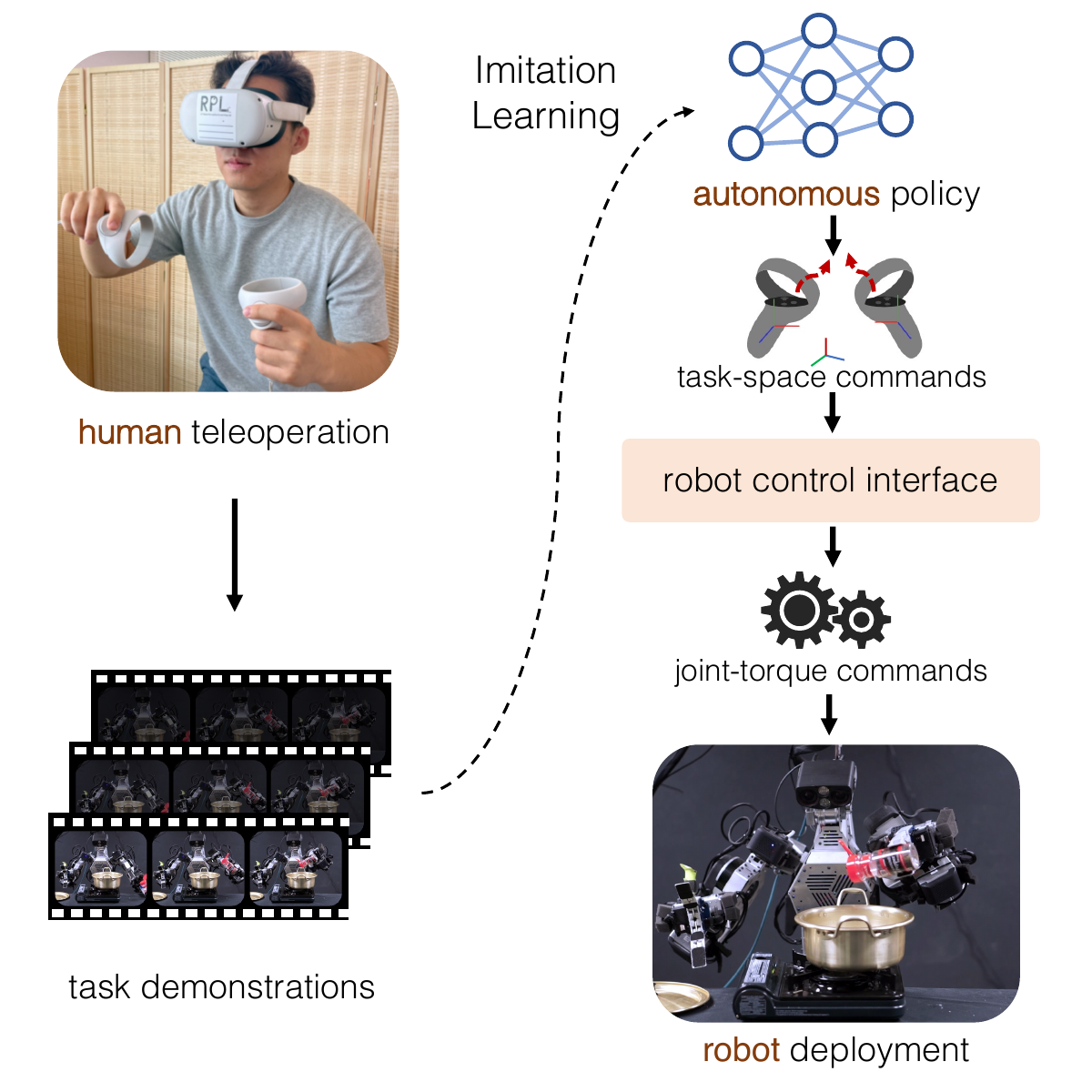}
	\caption
	{
        \textbf{Overview of \ourmethod{}.}
    \ourmethod{} addresses the challenge of learning humanoid loco-manipulation. 
    We introduce a learning framework that facilitates teleoperated demonstrations with task-space commands provided by a human demonstrator. 
    The trained policies leverage human complexity and adaptability in decision-making to generate these commands.
    The robot control interface then executes these target commands through joint-torque actuation while complying with robot dynamics.
    This synergistic combination of imitation learning and whole-body control enables successful method implementation in both simulated and real-world environments.
	}
	\label{fig:overview}
\end{figure} 

To this end, we introduce {\ourmethod{}} (\underline{T}eleope\underline{r}ation and \underline{I}mitation \underline{L}earning for \underline{L}oco-manipulation), a deep imitation learning framework for learning sensorimotor policies of humanoid robots from human demonstrations. 
TRILL consists of three main components.
The first is a VR-based teleoperation interface, offering an intuitive way for humans to provide task demonstrations.
The second is a whole-body controller, which reliably converts task-space trajectories of human demonstrations into joint-torque actions. The use of whole-body control prioritizes robot stability and tracks limb trajectories to generate dynamically feasible motions.
The final component is a data-efficient imitation learning algorithm for training loco-manipulation policies. 
Our policy predicts target setpoints for the robot's hands and sends commands prescribing gait sequence commands, enabling sample-efficient training with high-level action abstractions. These components together enable \ourmethod{} to perform complex loco-manipulation tasks while adeptly stabilizing uncertain robot dynamics.

We evaluate our approach in simulation and real hardware settings. In two simulated environments, \ourmethod{} achieves success rates of 96\% for free-space locomotion tasks, 80\% for manipulation tasks, and 92\% for loco-manipulation tasks. Across all tasks, our method outperforms state-of-the-art imitation learning baselines~\cite{mandlekar2021matters} by 28\% in success rate. We also deploy our method to a real-world humanoid robot, DRACO 3~\cite{bang2022control}, achieving an average success rate of 85\% in two contact-rich manipulation tasks. To the best of our knowledge, this work presents the first successful implementation of deep imitation learning for learning visuomotor policies of complex manipulation tasks on real-world humanoid systems.

%% file: related_work.tex
\section{Related Work}

\subsection{Loco-manipulation of Humanoid Robots}
Humanoid robots present unique challenges due to their discontinuous movement and the need to maintain balance during task execution. This is in contrast with wheeled mobile manipulators, which have continuous manifolds for locomotion and manipulation, making feasible motion straightforward~\cite{nagatani2002motion, welschehold2017learning, xia2020relmogen}. 
To address the challenges of controlling humanoids, recent research has explored kinodynamic whole-body solutions \cite{dai2014whole, jorgensen2020finding, vaz2020material, ahn2021versatile, calvert2022fast}. These methods, while promising, offer only task-specific solutions, lack generality, and demand significant computational power.

An alternative solution to whole-body control is human teleoperation, which aims at mitigating control complexity and improving the robot's interaction with the environment. Pioneering works~\cite{tachi2003telexistence, sian2004whole, stilman2008humanoid} have developed teleoperated robots and transferred simple human operator motions to humanoid robots at the whole-body level.
However, ensuring smooth, stable, real-time motions while maintaining the robot's balance remains a significant challenge when teleoperating highly dynamic motions. 
Recent works~\cite{jorgensen2019deploying, purushottam2023dynamic} have attempted to apply inverse dynamics approaches to handle the robot's changing dynamics, but these methods are computationally intensive and subject to numerical ill-conditioning.
Such limitations have narrowed the range of tasks that can be handled by teleoperation and have hindered the full implementation of learning from demonstration in the teleoperation of humanoid robots. 
Our work leverages the task-space action abstraction (as opposed to retargeting low-level joint motion) through a whole-body controller. It reduces computation complexity and improves the robustness in stabilizing the robot's dynamics.

\subsection{Imitation Learning from Teleoperation Demonstrations}
Learning from demonstration presents an effective approach to building robot behaviors with human supervision for complex and dexterous manipulation tasks \cite{xie2020deep, zhao2023learning}.
{Teleoperated human demonstrations have been shown to be particularly useful in reducing domain gaps between the training data and the deployment settings.}
Moreover, data collection can be scaled up with relative ease \cite{mandlekar2019scaling, mandlekar2021matters, tung2021learning, wong2022error}.
The benefits offered by this methodology have created broad interest in the robotics community, with extensive literature~\cite{mandlekar2021matters, nasiriany2022learning, zhu2022viola} exploring its potential in programming complex manipulation controllers where manual engineering is infeasible.
Most existing methods are constrained to tabletop manipulation or wheeled platforms. Unlike these platforms, collecting demonstrations for humanoids is significantly more challenging, as action commands from human operators cannot be easily mapped to the robots. This is primarily due to the complex floating-base dynamics of humanoid robots, compounded by stability issues and uncertainties in state estimation. As part of this work, we develop a practical system that enables large-scale collection of human demonstrations for humanoid loco-manipulation.









%% file: method.tex
\section{Method}

\begin{figure*}
	\centering
	\includegraphics[width=\linewidth]{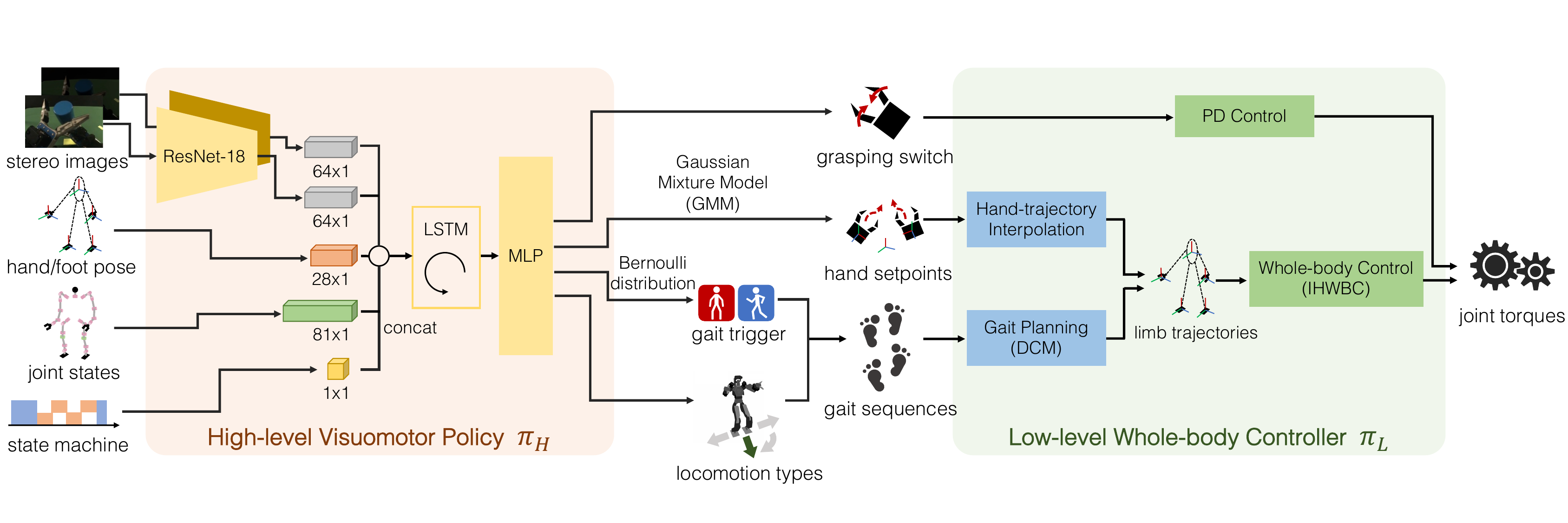}
	\caption
	{
        \textbf{Model architecture of \ourmethod{}.}
    The trained policies generate the target task-space command $u_t$ at 20 Hz from the onboard stereo camera observations and the robot's proprioceptive feedback. The robot control interface realizes the task-space commands, computes the desired joint torques $a_t$ at 100 Hz, and sends them to the robot for actuation.
    	}
	\label{fig:model}
\end{figure*}


We introduce \ourmethod{}, a deep imitation learning framework for humanoid loco-manipulation. The key to our approach is to decompose the loco-manipulation pipeline into a two-level hierarchy consisting of a high-level visuomotor policy and a low-level whole-body controller. The high-level action abstraction of the policy facilitates data-efficient learning. The low-level controller complements the high-level policy and stabilizes the robot dynamics while realizing the policy's intended actions.

\subsection{Problem Formulation}
We model the problem of loco-manipulation as a discrete-time Markov Decision Process $\mathcal{M}=(\mathcal{S}, \mathcal{A}, \mathcal{P}, R, \gamma, \rho_{0})$ where $\mathcal{S}$ is the state space, $\mathcal{A}$ is the action space, $\mathcal{P}(\cdot|s, a)$ is the transition probability, $R(s, a, s')$ is the reward function, $\gamma \in [0, 1)$ is the discount factor, and $\rho_{0}(\cdot)$ is the initial state distribution. Our goal is to learn a closed-loop visuomotor policy $\pi(a_t|s_t)$ that maximizes the expected return $\mathbb{E}[\sum^\infty_{t=0}\gamma^t R(s_t, a_t, s_{t+1})]$. In our context, $\mathcal{S}$ is the space of the robot's sensory observations captured by its egocentric cameras and proprioceptive sensors, $\mathcal{A}$ is the space of the robot's joint-torque commands,  $R(s, a, s')$ is the reward function designed for the loco-manipulation task, and $\pi$ is a closed-loop policy that we deploy on the robot. 

To handle the complexity of visuomotor skills and train the policy $\pi$ effectively, we decompose the policy $\pi$ into a two-level hierarchy. At the high level is a neural network policy $\pi_{H}$ that computes the target motion command $u$ as the hands' pose setpoints and the body's locomotion. At the low level is a whole-body controller $\pi_L$ that computes the robot's joint-torque actions to realize target commands $u$ from $\pi_H$. With this hierarchical abstraction, we can rewrite the whole policy as $\pi(a_t|s_t) = \pi_{L}(a_t|s_t, u_t)\pi_{H}(u_t|s_t) $.

\ourmethod{} utilizes this hierarchical structure, as illustrated in Figure~\ref{fig:model}.
The high-level policy $\pi_H$ generates task-space commands. We train $\pi_H$ through imitation learning from human demonstrations collected through our VR teleoperation system. The low-level control policy $\pi_L$ calculates motor torques to fulfill the commands set by $\pi_H$. We use the whole-body control formulation to implement the controller $\pi_L$ in order to ensure the robust execution of the robot's motions.


\begin{figure*}[t]
	\centering
	\includegraphics[width=\linewidth]{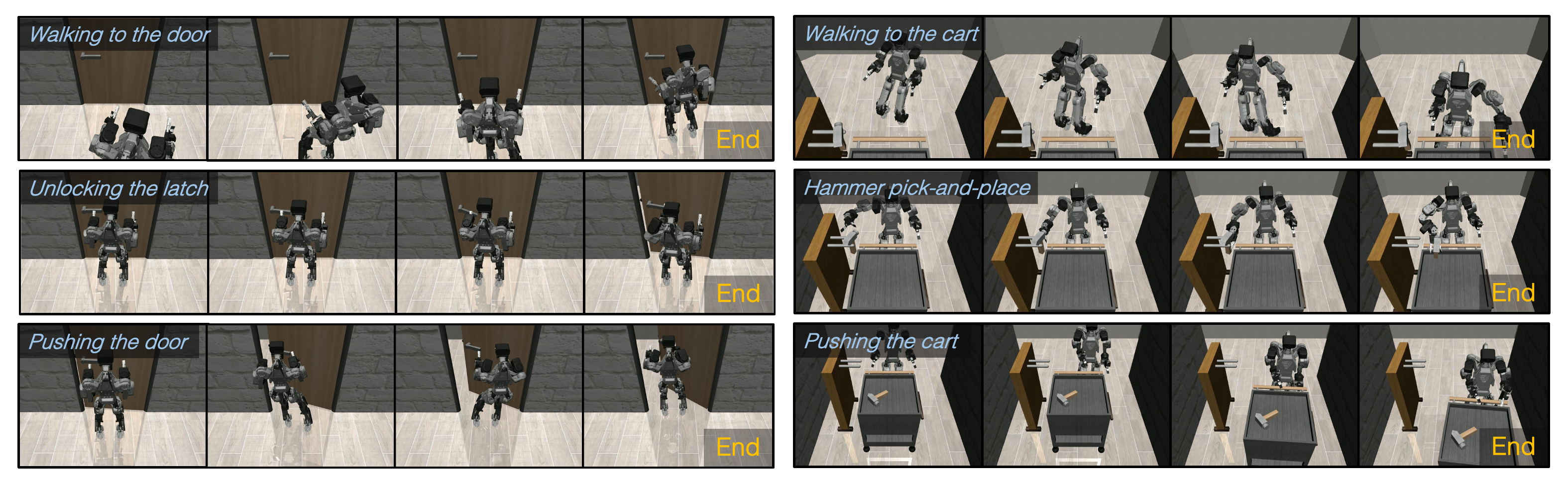}
	\caption
	{
        \textbf{Timelapse of deploying \ourmethod{} in simulation.}
        We present the deployment of policies trained through our method across three distinct tasks: free-space locomotion, manipulation, and loco-manipulation, in two unique simulation environment domains.
        }
	\label{fig:simulation}
\end{figure*}

\subsection{Teleoperation System}
\label{sec:gait}
We define the Cartesian hand poses in the robot's body frame as setpoints of robot arm motions. The poses of the VR hand controllers, with respect to the VR headset, are mapped to these setpoints. In this way, the task-space commands of hand poses reduce the action space and compensate for the lack of joint sensory feedback during teleoperation. We apply trajectory interpolation to prevent aggressive or discontinuous motion trajectories to ensure safe teleoperation. Further examination of the impact of this handling hand poses trajectories is detailed in Section \ref{sec:sim-eval}.

We model the locomotion commands as easily executable, predefined gait sequences to simplify our teleoperation operation. The buttons on the VR hand controllers trigger these sequences. Concretely, we employ the Divergent Component of Motion (DCM) approach for gait planning \cite{takenaka2009real}.
Based on the unstable portion of linear inverted pendulum dynamics, the DCM planner calculates the divergent component of motion. Unlike other methods that involve direct control of the robot's Center of Mass (COM), the DCM method allows for indirect COM control. It provides simplicity of the planner and a larger basin of attraction. These features are significantly beneficial in loco-manipulation, as they allow more effective control of hand positions relative to the robot's body motion. 

\subsection{Whole-body Control}
Although the morphological similarities between humans and humanoid robots make teleoperation more intuitive, the differences in kinodynamic properties can significantly complicate robots' low-level joint control.
The lack of proprioceptive feedback makes it further difficult for human demonstrators to stabilize the robot's pose.
Particularly, physical interactions (e.g., manipulating heavy objects) alter the robot's dynamics. With off-the-shelf VR controllers, human demonstrators must be able to operate the robot safely without force feedback from objects of unknown properties.

Given these challenges, a robust low-level whole-body motion controller is critical for effectively controlling humanoid robots. A desirable motion controller should fulfill two key requirements: 1) it should provide safe stabilization of the robot's body under external disturbances or underlying model inaccuracy, abstracting away the floating-base dynamics from human operators' input, and 2) it should have the capability to accurately track constantly varying trajectory commands, as set by $\pi_H$, to generate appropriate locomotion and manipulation behaviors for contact-rich tasks.

Inspired by recent advances in whole-body control \cite{koolen2016design, lee2021whole, ahn2021versatile, wensing2016improved}, we have employed a model-based optimal control method to calculate motor torques. Our controller stabilizes the robot's floating base against the uncertainties and disturbances caused by object interactions and maps closed-loop commands accurately into the robot's trajectories. Specifically, we have implemented the Implicit-hierarchical Whole-body Control (IHWBC) \cite{ahn2021versatile}. The formulation of the whole-body control is given as,
\begin{equation}
\label{eq:wbc}
\begin{aligned}
\min_{\ddot{q}, f_r}
 \sum_{i=1}^{n} 
  &w_i \|J_i\ddot{q} 
   +\dot{J}_i\dot{q}_m
   -\ddot{x}_i^d\|^2\\
  &+w_{f_r} \| f_r^d - f_r\|^2
  +\lambda_q \|\ddot{q}\|^2
  +\lambda_{f_r} \|f_r\|^2,\\
\text{subject to}\\
    &S_f(A\ddot{q} + b + g - J^\top f_r)=0\\
    &Uf_r \geq0\\
    &S_r f_r \leq f_r^\text{max}\\
    &\ddot{q}_\text{min} \leq \ddot{q} \leq \ddot{q}_\text{max}\\
    &\tau_\text{min} \leq S_a (A \ddot{q} + b + g - J_c^\top f_r) \leq \tau_\text{max}.
\end{aligned}
\end{equation}
The notation follows Section 3 in the original work of IHWBC \cite{ahn2021versatile}.
In this optimization formulation, $i$ denotes the terms used for controlling limb poses within IHWBC, such as the center of the robot's body, hands, and feet. By prioritizing body pose stabilization, the robot can tolerate minor tracking errors of its end-effectors while ensuring the body's stability. This prioritization prevents the robot from failing due to aggressive motion trajectories.
In our model, target reaction forces $f_r$ at the non-contact points on the hands and feet are processed as virtual impedances relative to target trajectories. This approach allows the robot to exhibit adaptive, compliant manipulation during object interactions.
Following Equation \eqref{eq:wbc}, the target command torque is determined as, 
$\tau^\text{cmd}=S_a(A\ddot{q}^*+b+g-J_c^\top f_r^*).$
Subsequently, the final joint torque inputs $\tau^\text{joint}=\tau^\text{cmd}+k_p(q^\text{cmd}-q_m)+k_d(\dot{q}^\text{cmd}-\dot{q}_m)$ are sent out to the robot.

\subsection{Training of Visuomotor Policies} 
\label{sec:policy}
Our collected demonstration dataset, denoted as $\mathcal{D}$, is comprised of state-action pairs $\{\mathcal{D}={(s_i, u_i)}_{i=1}^N\}$. Here, $N$ stands for the total number of data points.
The observation $s_i$ includes stereo images from the robot's on-board camera, actual positions of the robot's feet (14D) and hands (14D), the status of the controller's state machine (1D), and sine and cosine representations of joint positions (54D) and joint angular velocities (27D) for the arms, legs, and neck.
The demonstration commands $u_i$ include the subsequent setpoints for the hands (14D), grasping actions for both hands (2D), and locomotion instructions (1D).

We use this dataset to train our policy $\pi_H$ with a deep imitation learning algorithm. 
In particular, we train a behavioral cloning policy with recurrent neural networks (RNNs)~\cite{mandlekar2021matters} to capture the temporal dependencies of the robot's movements (see Figure~\ref{fig:model}).
To handle the inherently noisy and multimodal nature of human demonstrations, $\pi_H$ utilizes a Gaussian Mixture Model (GMM)~\cite{wang2020critic} for policy output.
We sample hand pose setpoints from the GMM and generate trajectories to these setpoints via interpolation.

The locomotion of the humanoid robot is discretized by abstracting the robot's walking modalities as predefined gait sequences. It simplifies the teleoperation system and allows the demonstrator to operate the robot's locomotion with button clicks. 
To improve the data efficiency of learning locomotion actions, we introduce a hierarchical decomposition of the action spaces into two components: \textit{gait trigger} and \textit{locomotion types}. The \textit{gait trigger}, sampled from a Bernoulli distribution whose probability distribution is dictated by the outputs of the high-level policy, determines if the gait is activated. Once the gait is activated, the values of the \textit{locomotion types} determine the final gait sequence types and produce the corresponding gait commands. Such factorization of the action space enables more sample-efficient training for locomotion, especially given the relatively sparse locomotion commands in the demonstration dataset.

%% file: experiments.tex
\section{Experiments}
\label{sec:exps}

In this section, we demonstrate the feasibility and effectiveness of \ourmethod{} for data collection and policy deployment on a humanoid robot, both in simulated and real-world settings. Furthermore, taking advantage of the scalability and ease of simulation environments, we use them to investigate the following research questions:
1) What factors impact humans' teleoperation performance and the ease of use of our VR-based interface?
2) How do different choices of observation and action spaces affect policy performances?
3) How does the policy performance scale with respect to the demonstration dataset sizes?

\subsection{Experimental Setup}
\label{sec:exp-setup}

We evaluate our method in a variety of locomotion and manipulation tasks.
A task is considered successful if the robot accomplishes the designated goals within a specific time limit without falling or experiencing undesirable collisions with objects. 
If the robot accomplishes the objectives but has undesired collisions with objects or only partially completes the tasks, we consider the task partially successful. 
To thoroughly evaluate our model's performance, we construct diverse task environments for simulation and real-world deployment with the DRACO 3 humanoid robot \cite{bang2022control}.

\paragraph{Simulation Setup}
In simulation, we design two realistic benchmark environments to study the robot’s loco-manipulation abilities: {\it Door} and {\it Workbench} in MuJoCo \cite{todorov2012mujoco}, as shown in Figure \ref{fig:simulation}. In each environment, we evaluate the robot’s ability to successfully perform subtasks involving free-space locomotion, manipulation, and loco-manipulation. 

In the {\it Door} environment, the robot performs the following three subtasks.
\begin{itemize}
    \item \task{Walking to the door} \textbf{(locomotion):} The robot walks toward a door and reaches its handle through free-space locomotion.
    \item \task{Unlocking the latch} \textbf{(manipulation):} The robot stands still and rotates the knob to unlock the door.
    \item \task{Pushing the door} \textbf{(loco-manipulation):} While walking, the robot pushes the door, thereby combining locomotion and manipulation. This task involves coordinated arm-base interaction, where the robot has to consider the mechanical constraints imposed by the door's mechanism.
\end{itemize}
Similarly, in the {\it Workbench} environment, the robot performs the following three subtasks:
\begin{itemize}
    \item \task{Walking to the cart} \textbf{(locomotion):} The robot approaches the cart. It requires precise positioning of the robot's body and its hands to avoid collision with objects.
    \item \task{Hammer pick-and-place} \textbf{(manipulation):} The robot lifts the hammer from the hanger and accurately places it onto the cart. In contrast to the \textit{Unlocking the latch} task in the \textit{Door} environment, this task demands precise dexterous skills and thus poses a greater challenge for manipulation.
    \item \task{Pushing the cart} \textbf{(loco-manipulation):} While walking towards the goal, the robot pushes a heavy cart. This task requires arm-base coordination, where the robot has to handle the planar motion constraints of the cart.
\end{itemize}

\paragraph{Real-World Setup}
Learning manipulation tasks on real-world robots presents additional challenges. This is primarily due to 1) the inherent uncertainty that arises from control and communication latency with real hardware systems, 2) the complex dynamics of robot actuators and the interactions of robots with objects in the real world, and 3) significant errors in estimating the robot's base state, hindering the precision of manipulation behaviors. To demonstrate the efficacy of \ourmethod{} given the floating-base dynamics of real humanoid robots, we have designed two intricate manipulation tasks. 
\begin{itemize}
    \item \task{Tool pick-and-place}\textbf{:} The robot is tasked with accurately locating and grasping a tool before placing it into a box. This task is designed to evaluate the robot's precision in handling manipulation tasks.
    \item \task{Removing a spray cap}\textbf{:} The robot is required to grasp a spray can with one hand and then remove its cap with the other hand. This task is designed to assess the robot's proficiency in bimanual manipulation.
\end{itemize}

\paragraph{Data Collection}
We adopt a VR device (Meta's Oculus Quest 2~\cite{meta2023}) for robot teleoperation. A stereo camera is mounted on the robot's head, with the stereo images used as visual observations for both the human operator and the visuomotor policy. The operator's hand poses are measured by the IMU on the Oculus hand controllers. Our experiments in simulation and with the real-robot system train policies on datasets of 250 demonstration trajectories for each task.

\begin{table}
\centering
\vspace{10pt}
\caption{\textbf{Quantitative results in simulation.} We report full and partial success rates (\%) on 25 trials of our \ourmethod{} policies against baselines. The numbers in parentheses are partial success rates.
}
\makeatletter\def\@captype{table}
\resizebox{\columnwidth}{!}
{
  \begin{tabular}{lcccc}
    \toprule
    \textbf{}  & \textbf{BC} \cite{finn2016deep}
               & \textbf{BC-RNN} \cite{mandlekar2021matters}
               & \textbf{TRILL} \\
    \midrule 
        \it{Walking to the door} 
        & 0 (0) & 40 (80)
        & {\bf 100} ({\bf 100}) \\
        \it{Unlocking the latch} 
        & 48 (52) & 80 (96)
        & {\bf 92} ({\bf 100}) \\
        \it{Pushing the door} 
        & 0 (0) & 56 (64)
        & {\bf 96} ({\bf 96}) \\
    \midrule
        \it{Walking to the cart} 
        & 0 (0) &  52 (88)
        & {\bf 92} ({\bf 96}) \\
        \it{Hammer pick-and-place} 
        & 0 (16) & 60 ({\bf 80})
        & {\bf 68} (76) \\
        \it{Pushing the cart} 
        & 0 (0) & 80 (\bf{88})
        & {\bf 88} ({\bf 88}) \\
    \bottomrule
  \end{tabular}
}
\label{tab:main}
\end{table}


\begin{table*}[t]
\centering
\vspace{10pt}
\caption{\textbf{Ablation studies of ~\ourmethod{} in simulation.} We use full and partial success rates (\%) on 25 trials of trained policies as our evaluation metric. The numbers in parentheses are partial success rates. In addition, we present a comparison of the success rate of our final model, with changes indicated by color. The red color signifies a decrease, while the blue color denotes an increase.
}
\makeatletter\def\@captype{table*}
\resizebox{1.7\columnwidth}{!}
{
  \begin{tabular}{lcccccc}
    \toprule
    \textbf{} & \task{monoscopic} & \task{no-joint-state} 
               & \task{deterministic} & \task{categorical}\\
    \midrule 
        \it{Walking to the door} 
        & 100 \same{} (100 \same{}) & \,\;92 \worse{8\,\;} (\,\;96 \worse{4\;\;})
        & \,\;92 \worse{8\,\;} (100 \same{}) & \,\;52 \worse{48} (\,\;84 \worse{16}) \\
        \it{Unlocking the latch} 
        & \,\;56 \worse{36} (\,\;84 \worse{16}) & \,\;64 \worse{28} (\,\;84 \worse{16})
        & \,\;40 \worse{52} (\,\;72 \worse{28}) & \,\;92 \same{} (\,\;96 \worse{4\,\;}) \\
        \it{Pushing the door} 
        & \,\;16 \worse{80} (\,\;32 \worse{64}) & \,\;24 \worse{72} (\,\;64 \worse{32})
        & \,\;24 \worse{72} (\,\;28 \worse{68}) & \,\;80 \worse{16} (\,\;80 \worse{16}) \\
    \midrule
        \it{Walking to the cart} 
        & 100 \better{8\,\;} (100 \better{4\,\;}) & \,\;88 \worse{4\,\;} (100 \better{4\,\;})
        & \,\;36 \worse{56} (\,\;80 \worse{16}) & \,\;92 \same{} (\,\;96 \same{}) \\
        \it{Hammer pick-and-place} 
        & \,\;28 \worse{40} (\,\;48 \worse{28}) & \,\;12 \worse{56} (\,\;20 \worse{56})
        & \,\;\,\;0 \worse{68} (\,\;\,\;0 \worse{76}) & \,\;36 \worse{32} (\,\;64 \worse{12}) \\
        \it{Pushing the cart} 
        & \,\;24 \worse{64} (\,\;44 \worse{44}) & \,\;52 \worse{36} (\,\;60 \worse{28})
        & \,\;\,\;4 \worse{84} (\,\;\,\;4 \worse{84}) & \,\;32 \worse{56} (\,\;40 \worse{48}) \\
    \bottomrule
  \end{tabular}
}
\label{tab:ablation}
\end{table*}


\subsection{Quantitative Evaluation in Simulation}
\label{sec:sim-eval}

\begin{figure*}[t]
	\centering
	\includegraphics[width=\linewidth]{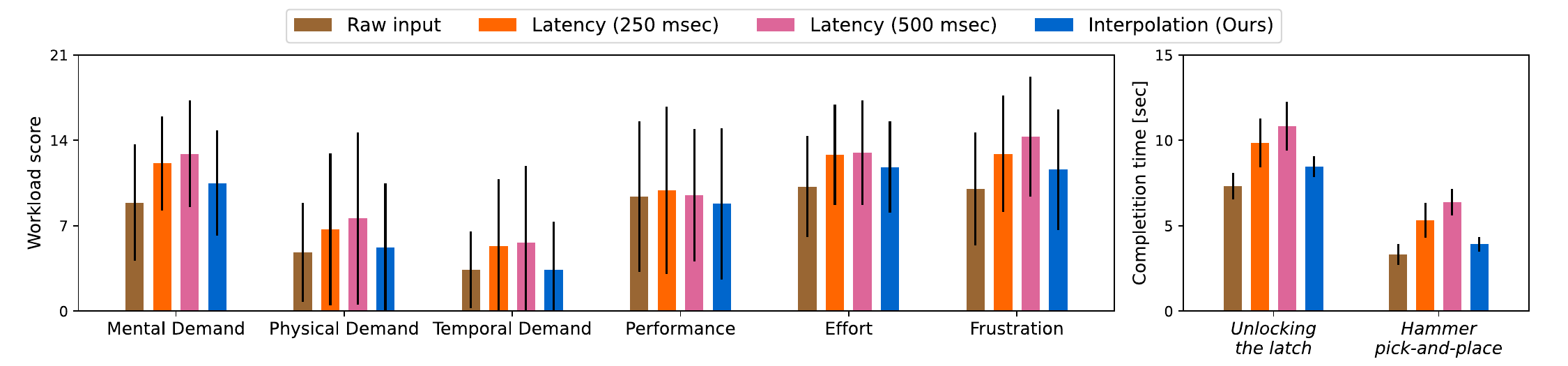}
	\caption
	{
        \textbf{NASA TLX evaluation and completion times using the teleoperation system:} We evaluate our teleoperation system using participants' self-reports from the TLX (left) and completion times on 25 trials (right) for two manipulation tasks in simulation. Lower scores indicate better performance in both categories. We further examine the effect of varying teleoperation settings, specifically control latency and trajectory handling, on demonstrator performance.
        In our interface, the added workload from trajectory interpolation has a minimal effect on teleoperation performance, allowing our method to produce high-quality demonstration data without compromising robot safety.
	}
	\label{fig:interface}
\end{figure*}

\begin{figure}
	\centering
	\includegraphics[width=\linewidth]{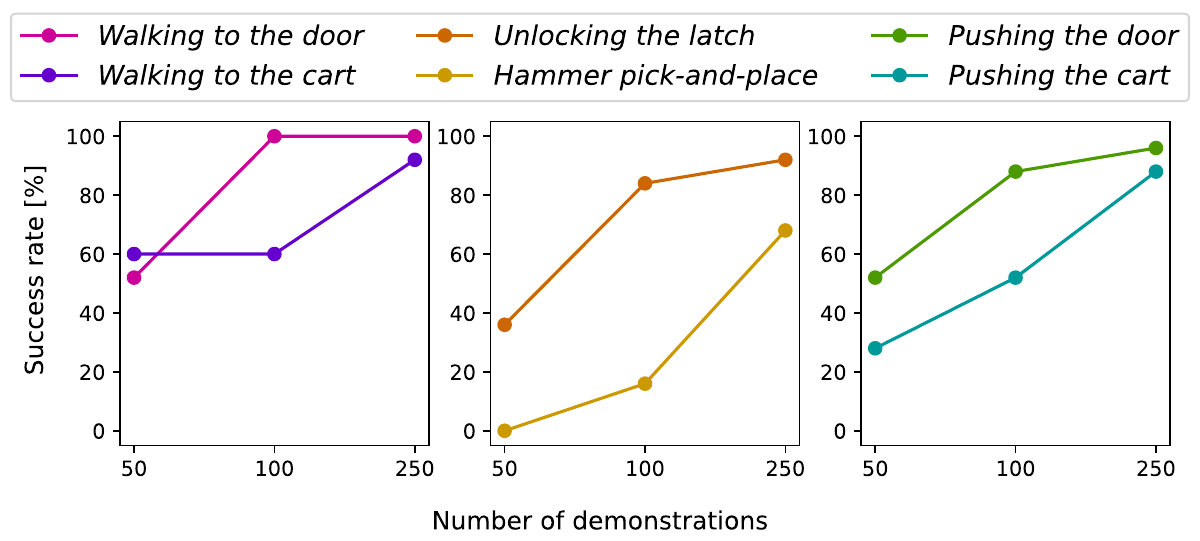}
	\caption
	{
        \textbf{Simulation evaluation across different dataset sizes.}
        We show the percentage changes in success rates on 25 trials, which were obtained from policies trained on datasets of different sizes. We observed that trained policies perform better across different tasks with an increase in demonstrations.
        }
	\label{fig:scale}
\end{figure}

\begin{figure*}[t!]
    \vspace{8pt}
    \centering
	\includegraphics[width=0.9\textwidth]{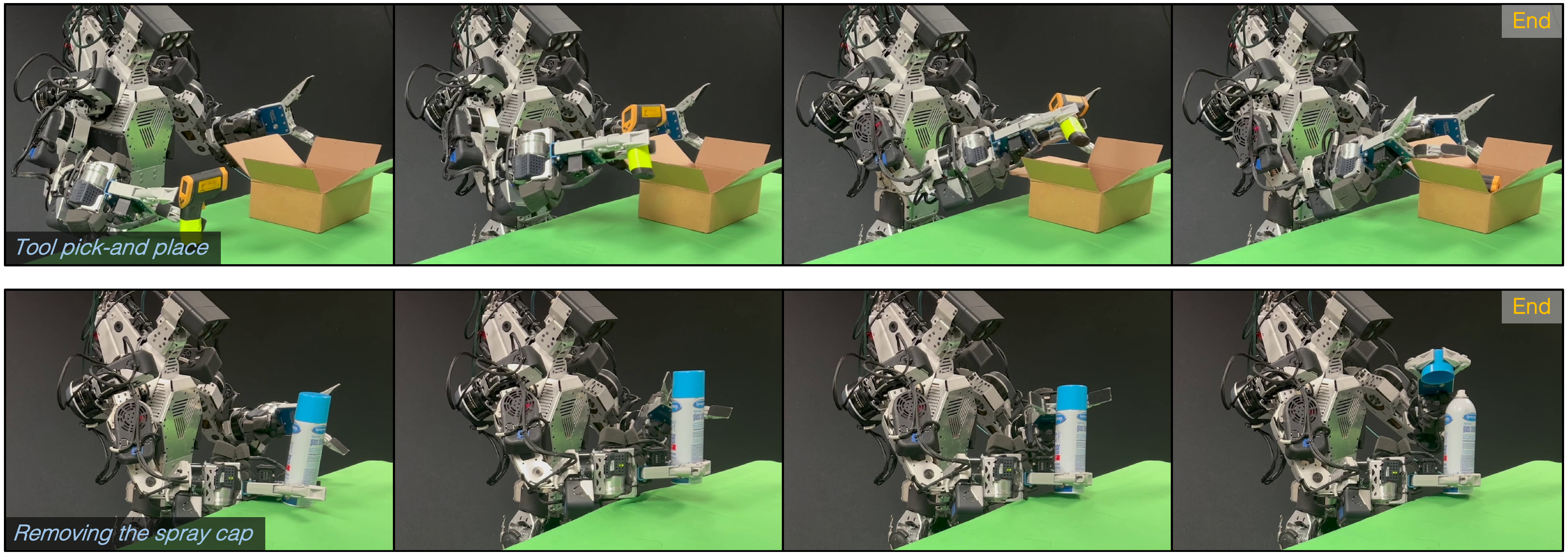}
	\caption
	{
        \textbf{Timelapse of deploying \ourmethod{} in the real DRACO 3 humanoid robot.}
        We deploy our method to a real humanoid robot to perform two bimanual manipulation tasks. The trained policies perform the \textit{Tool pick-and-place} task successfully in 8 out of 10 trials and the \textit{Removing the spray cap} task in 9 out of 10 trials, consecutively, without the need for rebooting the robot.
	}
	\label{fig:hardware}
\end{figure*}

\paragraph{Deployment Results} 
Table \ref{tab:main} reports quantitative evaluations of our simulated tasks. We compare our model's performance against the following baselines.
\begin{itemize}
    \item \textbf{BC:} a behavioral cloning baseline that learns a reactive policy conditioned on current observations. The model architecture is based on the design of Finn et al. \cite{finn2016deep}, which uses convolutional image encoders.
    \item \textbf{BC-RNN:} the state-of-the-art method that employs a temporal sequence of past observations using recurrent neural networks introduced in Mandlekar et al. \cite{mandlekar2021matters} for learning manipulation skills from teleoperated demonstrations. In contrast to the hierarchical action space used in \ourmethod{}, this baseline uses flat GMM outputs for both manipulation and locomotion commands.
\end{itemize}
As shown in Table \ref{tab:main}, \ourmethod{} outperforms the two baseline methods across all tasks in full success rates and five out of the six in partial success rates. The performance gaps are more significant in free-space locomotion and loco-manipulation tasks.
The BC-RNN baseline shows consistently better performances than BC, highlighting the importance of modeling the temporal dependencies for humanoid loco-manipulation skills. While these two baselines can learn policies that effectively produce hand trajectories for manipulation in some cases, they struggle to learn from the sparse locomotion commands in human demonstrations. 
Our model, which suitably formulates locomotion actions in separation with manipulation actions, can train visuomotor policies more efficiently than these baselines.

\paragraph{Teleoperation Interface Evaluation} 
We evaluate the intuitiveness of our VR-based teleoperation system. We also explore how the following factors influence user experience and demonstrator performance: 1) control delay and 2) {trajectory interpolation used in our teleoperation system.}
To assess these factors, we recorded 25 episodes of human demonstration for each of the two simulated manipulation tasks. We utilize the NASA Task Load Index (TLX) \cite{hart1988development} to quantitatively analyze workload and compare task completion time as the performance metrics.
\begin{itemize}
    \item \textbf{Raw input:} This baseline directly applies teleoperation commands without any trajectory interpolation. It is impractical in real-robot settings, as non-smooth trajectories caused by communication delays or unsafe teleoperation actions may cause damage. Nonetheless, this baseline allows us to methodically examine how our final setup's trajectory interpolation, as described in Section \ref{sec:policy}, affects performance.
    \item \textbf{Latency (250 msec):} This baseline is a variant of our final setup where a $250 \, \text{msec}$ latency is artificially injected in order to study the impact of control delays on demonstrators' performance.
    \item \textbf{Latency (500 msec):} Similarly to the Latency (250 msec) baseline, it features a significantly larger control delay that could potentially hinder the demonstrators' ability to perform tasks.
\end{itemize}
As indicated in Figure \ref{fig:interface}, we observe that as control latency rises, there is a corresponding increase in workload and a decrease in performance. It is also worth noting that this increase in workload burden impacts the demonstrator, consequently affecting the quality of the demonstration.

In our teleoperation setup, we implemented trajectory interpolation rather than directly using raw hand pose setpoints for limb trajectories at the whole-body controller. 
This was done to ensure safer robot operation, as well as smoother and less aggressive changes in command setpoints.  
When employing trajectory interpolation, there is only a minor impact on demonstration performance when compared to the Raw input baseline. 
It suggests that the additional workload imposed by trajectory interpolation does not significantly affect teleoperation performance. 
As a result, our teleoperation approach can produce high-quality demonstration data while maintaining safe robot operation.

\paragraph{Varying Dataset Sizes} 
We also conducted a systematic investigation into the effect of dataset size on trained policies by varying the dataset size between 50 and 250 demonstrations.
The results of this investigation are presented in Figure \ref{fig:scale}.
Our findings indicate that an increase in the number of demonstrations generally leads to improved performance of trained policies across various tasks regardless of their difficulty.

\paragraph{Ablation Studies} 
We conducted a study on the effect of observation and action space designs. In order to perform ablation studies, we compared the performance of model variants that employ different observation and action space designs, as described below.
\begin{itemize}
    \item \task{monoscopic}\textbf{:} In this model variant, only the right side of camera observation is used as input to the policy. This is intended to examine whether depth information derived from stereo images is essential for the robot's performance.
    \item \task{no-joint-state}\textbf{:} This variant excludes joint-state feedback in its observations. It is intended to examine the role of joint sensor feedback in learning locomotion and manipulation skills.
    \item \task{deterministic}\textbf{:} Instead of sampling output hand setpoints from a GMM, this variant adopts a deterministic action output. It examines the importance of modeling multi-modality in action distributions and how it helps stabilize humanoid robots over the errors induced by their floating dynamics and uncertainty.
    \item \task{categorical}\textbf{:} This variant employs a categorical action head to predict output locomotion commands. It is intended to assess the efficiency of our choice of action space tailored for humanoid locomotion.
\end{itemize}

We report the average success rate and partial success rate of each model variant in Table \ref{tab:ablation}.
Our findings highlight the vital role of stereoscopic visual observations in tasks requiring precise hand movements. We hypothesize that this is because stereoscopic visual observations provide key spatial information to resolve depth ambiguity. 
Although joint state sensor modalities do not show as big an impact as stereoscopic vision, they remain important for successful manipulation.
The \textit{deterministic} variant yields the most significant decline in success rates across manipulation and loco-manipulation tasks. It indicates that deterministic actions cannot effectively capture the action distribution's multi-modality given the complex robot dynamics.
Lastly, the \textit{categorical} variant showed the most considerable performance reduction in free-space locomotion and loco-manipulation tasks. It implies that separating locomotion actions into \textit{gait trigger} and \textit{locomotion types} improves learning efficiency for locomotion actions.

\subsection{Real-Robot Deployment}

\paragraph{Teleoperation Interface Evaluation} 
We present the tracking errors of our teleoperation interface. These errors primarily result from phase lags, tracking issues, and workspace limitations imposed for safety reasons. Hand position errors account for $0.082 \pm 0.032 \, \text{m}$ and hand orientation errors account for $22.2 \pm 10.9 \, ^{\circ}$ across human demonstrations.
Note that real robot systems, unlike idealized simulation environments, are subject to control latency. We previously discussed the effect of this latency on demonstrator performance in the simulation settings. On average, the control latency with the real system amounts to $122 \pm 190 \, \text{msec}$.
Moreover, error distribution in teleportation and control latency underlines the importance of data collection and policy deployment within the same settings. The embodiment-specific error and latency distributions pose significant challenges when transferring policies across different hardware platforms, from demonstration to deployment.

\paragraph{Deployment of the Trained Policies} 
Finally, we deploy our policies trained on real-world demonstrations to physical hardware. We aim to validate our policies' robustness against variations due to controller limitations, sensor inaccuracies, and floating-base dynamics. 
We focus on the bimanual manipulation tasks described in Section \ref{sec:exp-setup}.
During evaluation, the robot performed each task 10 times in a row without rebooting. \ourmethod{} succeeded in 8 out of 10 trials in the {\it Tool pick-and-place} task and 9 out of 10 trials in the {\it Removing the spray cap} task, respectively (see Figure \ref{fig:hardware}). More videos are provided on our project website.

%% file: conclusion.tex
\section{Conclusion}
\label{sec:conclusion}

We introduce TRILL, an effective imitation learning framework for teaching humanoid robots loco-manipulation skills. 
We developed an intuitive VR teleoperation system for human operators to supply demonstration trajectories at ease.
Through the use of whole-body control, our hierarchical approach robustly executes learned task-space commands while maintaining the dynamical stability of the humanoid robots.
We introduced a data-efficient deep imitation learning algorithm to train the high-level visuomotor policies. Our experiments have validated the effectiveness of our learning framework in physical simulation and on real-world hardware.
In the future, we aim to expand our framework to solve long-horizon tasks and further improve the generalization ability of the learned skills.

%% file: appendix.tex
\section* {Appendix}
\label{sec:appendix}

\renewcommand\thesubsection{\Alph{subsection}}

\subsection{Implementation Details}
The visuomotor policies generate the target task-space commands at 20 Hz. The robot control interface, including hand-trajectory interpolation, gait planning, and whole-body control, is updated and computes the joint-space commands at 100 Hz to actuate the robot. We provide implementation details for reproducibility.

\paragraph{Whole-body Control} 
In the whole-body control, we control 26 joints of the robot: 6 joints in each arm and 7 joints in each leg. Due to the knees being designed as rolling contact mechanisms, the two knee joints in each leg are interlinked. Consequently, the whole-body controller computes motor torques for 24 degrees of freedom (DOFs). Additionally, 1 DOF for each gripper is controlled through PD control, which operates outside the whole-body control. Likewise, the pitch joint in the robot's neck is controlled by PD control, and target joint angles are consistent throughout the environments.

In the prioritization scheme of IHWBC, maintaining body pose stability is always the top priority to ensure safe operation. This is followed by tracking foot pose trajectories and hand pose trajectories, respectively. When the robot is balancing on its two feet in contact with the ground without walking, the priority for tracking hand poses is increased to enhance manipulation performance. However, while the robot is walking, the priority of tracking hand poses is lowered to reduce the impact of hand motions on the robot's overall stability. The prioritization of stabilizing body poses and tracking foot trajectories remains consistent throughout trials.

\paragraph{Teleoperation Systems}
Hand pose setpoints in our teleoperation system are updated at 20 Hz. Smoothing between successive setpoints is achieved through trajectory interpolation, specifically using Hermite curves. For the locomotion commands, we employ the following pre-defined locomotion types:
\begin{itemize}
\item \task{forward}\textbf{:} The robot moves $0.2 \, \text{m}$ forward.
\item \task{backward}\textbf{:} The robot moves $0.2 \, \text{m}$ backward.
\item \task{left-turn}\textbf{:} The robot rotates its body $18 \, ^{\circ}$ in the left direction.
\item \task{right-turn}\textbf{:} The robot rotates its body $18 \, ^{\circ}$ in the right direction.
\item \task{left-sidewalk}\textbf{:} The robot moves $0.1 \, \text{m}$ to the left side without rotating its body.
\item \task{right-sidewalk}\textbf{:} The robot moves $0.1 \, \text{m}$ to the right side without rotating its body.
\end{itemize}
Upon receiving locomotion commands, gait trajectories are generated by the DCM planner, and the robot then executes them. The teleoperation system is designed to accept new locomotion commands only after the current gait sequence has been fully completed.

\paragraph{Visuomotor Policy Architecture} 

The controller's state machine assigns discrete values to track the robot's walking phases. These include the initiation and termination of ground contact for each leg, the swinging phase for each leg, and the balanced state when both feet are on the ground. The values of the state machine are essential for the visuomotor policy to effectively handle the robot's locomotion states. The robot's hand and foot positions are provided in Cartesian coordinates and quaternions in the robot's body frame. Joint positions are encoded using concatenated vectors of their sine and cosine values. The RGB images used as inputs are $400 \times 300$ pixels. The visuomotor policy employs two separate image encoders based on the ResNet18 architecture~\cite{he2016deep}. These encoders share the same architecture and are both trained end-to-end.

After the image features are encoded, they are flattened and concatenated with the data representing the robot's hand and foot poses, joint states, and the state machine value. This combined vector is then processed by recurrent neural networks. For the RNNs, we use Long Short-Term Memory (LSTM) networks~\cite{hochreiter1997long} of two layers with 400 hidden units for each layer. Finally, the policy outputs are delivered through a two-layer Multi-Layer Perceptron (MLP), with each layer containing 1024 hidden units. The GMM policy output has 5 modes.


For both the GMM and the Bernoulli distribution, the policy outputs the distribution parameters. Using the output of the GMM, we determine the next target pose for each hand by calculating the differences in Cartesian coordinates and quaternions from the frame of the previous hand pose. For the locomotion commands, a binary \textit{gait trigger}, sampled from a Bernoulli distribution, decides whether to commence the robot's walking.  When the \textit{gait trigger} is activated, the robot plans its gait trajectories according to the \textit{locomotion types} output by the policy. During the execution of these gait sequences, the robot disregards any new locomotion commands until the sequence is complete. After completion, it can accept new commands. 

For imitation learning, we employ behavioral cloning. We use the cross-entropy loss for action losses associated with grasping and the \textit{locomotion types}, as they are discrete outputs. For sampling of hand setpoints and the \textit{gait trigger}, we apply the negative log-likelihood loss for the probability distributions.

\subsection{Experimental Setup}
In both simulation and real-world environments, the same objects are used across the trials, but their initial locations and orientations are randomized. For the free-space locomotion tasks, the initial robot poses are varied within a large distribution. In contrast, for the loco-manipulation and manipulation tasks, while the robot poses are still randomized, the variation is limited to a smaller distribution.

Regarding the baselines of the Deployment Results in Section \ref{sec:sim-eval}, the observation inputs for the visuomotor policies remain consistent across the baselines, and the architecture of each baseline is suitably adapted. For the self-variants of the Ablation Studies in Section \ref{sec:sim-eval}, the observation inputs and policy outputs are consistent unless otherwise specified. Similarly, the architecture of each self-variant is modified correspondingly based on TRILL's architecture.